%% file: WhitePaper.tex
\documentclass{article}
\usepackage[preprint,eandd,nonatbib]{neurips_2026}
\usepackage[T1]{fontenc}
\usepackage{cite}
\usepackage{amsmath,amssymb,amsfonts}
\usepackage{graphicx}
\usepackage{textcomp}
\usepackage{xcolor}
\usepackage{booktabs}
\usepackage{array}
\usepackage{placeins}
\usepackage{url}
\usepackage[hidelinks]{hyperref}
\input{numbers.tex}
\input{eval/injecagent-numbers.tex}

\title{The Agent Incident Registry: Toward Preventing Repeated AI Agent Failures}
\author{%
Divyanshu Kumar \\
Anaconda \\
\texttt{dkumar@anaconda.com}
\And
Rohith HN \\
Anaconda \\
\texttt{rhn@anaconda.com}
\And
Nitin Aravind Birur \\
Anaconda \\
\texttt{nbirur@anaconda.com}
\And
Sahil Agarwal \\
Anaconda \\
\texttt{sagarwal@anaconda.com}
\And
Prashanth Harshangi \\
Anaconda \\
\texttt{pharshangi@anaconda.com}
}

\begin{document}
\maketitle

\begin{abstract}
AI agents increasingly act through tools and delegated authority, but general incident
repositories rarely capture the mechanisms needed to compare public failures with
agent-security evaluations. We present the Agent Incident Registry (AIR)\footnote{Project page: \url{https://enkryptai.com/air}.}, a source-linked
catalog containing 487 records of agent-related events disclosed from 2022 through
2026. Each record includes supporting evidence, a stable identifier, and
missingness-aware labels for causal role, disclosure class, mechanism, and outcome. Among
the 336 generative-system records in which the agent acted, 81 involved
realized harm (24\%). Realized outcomes concentrate in in-the-wild and safety-failure
records, while responsible disclosures and research demonstrations are overwhelmingly
demonstrated; the aggregate share therefore characterizes collection composition rather than
deployment risk. After initial curation, a second human reviewer checked all 487 records and
their existing labels for completeness and correctness. In a deployment-analogue audit,
InjecAgent's 1,054 cases occupy three of AIR's twelve surfaces and are all
attacker-triggered, whereas AIR contains 92 no-adversary safety failures. AIR supports
source-grounded case retrieval and evaluation-scope auditing, not failure-rate or
control-efficacy estimation.
\end{abstract}


\section{Introduction}

In July 2025 a coding agent on Replit reportedly deleted a live production database during an
active code freeze despite repeated instructions not to make changes
(\texttt{AIR-2025-0061}).
In June 2025 researchers demonstrated that a crafted email could make Microsoft 365 Copilot
exfiltrate data without the user clicking the malicious content, although retrieval occurred
during a user-initiated Copilot interaction. Despite the cited paper's ``real-world exploit''
title, AIR codes this event as demonstrated rather than realized because no customer harm was
documented (\texttt{AIR-2025-0039}; \cite{reddy2026echoleak}). One event caused real data
loss; the other established a feasible exploit. Treating both as undifferentiated
``incidents'' obscures a distinction that evaluation designers need.

Agent failures are system failures: model behavior combines with credentials, tool access,
untrusted content, and delegated authority to produce consequences. General AI incident
repositories capture a broader range of harms, but their public schemas do not consistently
record these agent-specific mechanisms. As a result, they cannot systematically connect public
events to the threat models and environments used in agent-security evaluations.

AIR addresses this gap with \N{} source-linked registry records representing events disclosed
between \Yfirst{} and \Ylast{}. Every admitted record has a supporting source and verbatim
evidence quote, a stable \texttt{AIR-YYYY-NNNN} identifier, and explicit missingness for
fields that the source does not establish. The schema separates \emph{realized harm}, in
which a real party experienced a consequence, from \emph{demonstrated capability}, and
distinguishes who acted from how the event reached disclosure.

\paragraph{Contributions.}
\begin{itemize}
  \item \textbf{A source-grounded registry.} AIR consolidates \N{} public disclosures into
  deduplicated records. Each record preserves a supporting quotation and source, a stable
  identifier, and explicit unknowns; documented merge rules prevent multiple reports of one
  event from being counted as separate incidents.
  \item \textbf{A mechanism-focused coding scheme.} AIR separates causal role, disclosure
  class, and realized outcome, so a demonstrated vulnerability is not counted as field harm
  and an event involving AI is not automatically attributed to agent action. A second human
  reviewer checked every record and existing label across the full catalog.
\end{itemize}

Together, the registry and coding scheme make public disclosures usable as evidence-grounded
inputs to agent-security evaluation: they show which mechanisms have reached disclosure and
which documented failure modes constructed tests may omit. Throughout the paper, AIR is
treated as a selected corpus of public disclosures, not a census of deployed systems or agent
runs; it therefore supplies neither deployment denominators nor counterfactual systems for
estimating failure rates or causal effects. Section~\ref{sec:method} defines its scope and
evidence rules, and Section~\ref{sec:threats} sets the limits on statistical interpretation.

\section{Related work}
\paragraph{Incident repositories.}
The AI Incident Database (AIID) established the case for learning from deployed AI failures and
groups public reports into incident records \cite{mcgregor2021aiid}. Its optional CSET
taxonomy distinguishes realized and potential harm and includes an autonomy judgment
\cite{hoffmann2023cset}; GMF relates goals, methods, and known or potential failure causes
with evidence-grounded rationales \cite{pittaras2023gmf}. The OECD subsequently defined
incidents and hazards \cite{oecd2024definitions}, proposed a common reporting framework
\cite{oecd2025reporting}, and built AIM, which explicitly includes both incidents and hazards
and generates harm, severity, stakeholder, and geography metadata
\cite{oecd2026aim}. AIAAIC manually curates incidents and controversies across AI,
algorithms, and automation \cite{aiaaic2026}, while the MIT AI Incident Tracker re-annotates
AIID records using causal and domain taxonomies
\cite{slattery2026repository,mit2026tracker}.

These resources cover a broader universe than AIR, and several already encode concepts that
AIR uses. The narrower distinction is that, at our freeze, their documented core schemas did
not consistently require the combination of agent role, autonomy, tool capability, initial
vector, guardrail outcome, and source evidence. AIR makes that agent-specific mechanism
schema mandatory while representing unsupported fields as missing. Table~\ref{tab:related}
summarizes this distinction.

\begin{table}[t]
\caption{AIR complements general AI incident repositories with required agent-mechanism
fields. ``Not explicit'' means the resource may describe a fact in prose or an optional
taxonomy but does not require it in its core documented schema.}
\label{tab:related}
\centering
\small
\setlength{\tabcolsep}{3.5pt}
\begin{tabular}{@{}p{0.16\linewidth}p{0.25\linewidth}p{0.31\linewidth}p{0.19\linewidth}@{}}
\toprule
Resource & Unit / principal source & Principal coding & Agent mechanism \\
\midrule
AIID & Incident linked to public reports & Harms; optional CSET/GMF taxonomies & Partial: optional autonomy/causes \\
OECD AIM & Incident or hazard; clustered news & Harm, severity, stakeholders, geography & Not explicit \\
AIAAIC & Incident or controversy; public sources & Risk, harm, sector, system and governance & Not explicit \\
MIT tracker & Re-annotated AIID incident & Causal/domain risk and harm severity & Not explicit \\
AIR & Registry record linked to public sources & Architecture, mechanism, control, agency, outcome & Explicit \\
\bottomrule
\end{tabular}
\end{table}

Cybersecurity and aviation provide methodological precedents for this narrower claim. VCDB
publishes disclosed breaches while warning that legal requirements skew its source population
\cite{veris2026vcdb}. NASA's voluntary ASRS fuses multiple reports into unique incidents and
states that its reports cannot estimate total frequency or a stable trend
\cite{nasa1996asrs}. NVD audits likewise show that catalog inconsistencies can change
downstream conclusions \cite{anwar2021nvd}. AIR adopts the same discipline: preserve
source-linked evidence, define the counting unit, and avoid prevalence claims.

\paragraph{Taxonomies and identifiers.}
OWASP ASI and AIVSS, MITRE ATLAS, and NIST's risk and adversarial-ML frameworks organize
risks, vulnerabilities, controls, and techniques rather than disclosed events
\cite{owasp2025agentic,owasp2026aivss,mitre2026atlas,tabassi2023airmf,vassilev2025aml}.
AIR maps every in-scope record to one or more OWASP ASI mechanisms and explicitly marks records
outside that taxonomy's scope. ATLAS and CVE links remain optional; a missing link is not
treated as evidence of absence. AIR's immutable identifiers follow the separation between
records and classifications used by CVE and CWE \cite{mitre2026cve,mitre2026cwe}.

\paragraph{Agent evaluations.}
These taxonomies name mechanisms; agent evaluations instantiate them as executable tasks.
Indirect prompt injection established that untrusted retrieved content can redirect
LLM-integrated applications without direct access to the model \cite{abdelnabi2023indirect}.
AgentDojo operationalizes that threat in stateful tool environments: formal utility and security
checks inspect resulting environment state across 97 benign tasks and 629 security cases, so it
tests tool-mediated consequences rather than text compliance alone
\cite{debenedetti2024agentdojo}. InjecAgent contributes a larger Cartesian suite of indirect
injections across user and attacker tools \cite{zhan2024injecagent}. Agent Security Bench
broadens the attack and defense families across ten scenarios and more than 400 tools
\cite{zhang2025asb}, while AgentHarm tests whether agents refuse harmful multi-step tasks while
retaining benign capability \cite{andriushchenko2024agentharm}.

Taxonomy-driven black-box red teaming offers a complementary route: a seven-domain framework
uses SAGE-RT \cite{kumar2024sagert} to generate adversarial scenarios and evaluate agents
without privileged internal access \cite{kumar2026blackbox}. AIR is being used to ground this workflow in disclosed
incidents: records provide evidence-backed scenario seeds, while its mechanism fields expose
gaps in generated-suite coverage.

These benchmarks and generated suites answer whether an agent fails under a constructed task
and threat model.
AgentDojo is especially close to AIR's tool-mediated mechanism level, but its security cases
remain synthetic adversarial demonstrations; AgentHarm covers harmful use, not spontaneous
no-adversary failure. AIR instead describes selected public events. It can audit which observed
surfaces, vectors, consequence types, and no-adversary cells an evaluation represents. We
perform one pinned InjecAgent projection in Section~\ref{sec:findings} as a worked example.
Applying that comparison requires explicit rules for what enters AIR and what one registry
record represents.

\section{Method}
\label{sec:method}

\subsection{Scope, unit, and populations}

\paragraph{Scope.}
AIR includes disclosures in which a generative or physically embodied autonomous system acts,
becomes the target of an AI-specific exposure, produces output on which a person acts, or
undergoes a bounded agent-security test. Included records must involve at least one
agent-specific mechanism, such as tool use, retrieval, delegated authority, autonomous
control, AI-specific data handling, or an agent-facing integration. We exclude generic
text-only jailbreaks without a concrete security or safety-control failure, deepfakes,
model-output bias audits, training-data disputes, and non-generative classifiers making
eligibility or ranking decisions.

\paragraph{Counting unit and identifiers.}
AIR's counting unit is a registry record, not an article, victim, repository, or download. A
record may represent one disclosed event, campaign, coordinated advisory, or distinct
vulnerability under the stated merge and split rules. Multiple reports of the same event are
fused, and a campaign with many artifacts still receives one record; proportions therefore
weight registry records rather than affected entities. A vulnerability disclosure and a later
exploitation campaign are split only when the later event has a distinct source and
realized-harm question. Each retained record receives an append-only
\texttt{AIR-YYYY-NNNN} identifier that is never reassigned.

\paragraph{Analysis populations.}
Causal role is recorded in four mutually exclusive \texttt{agency} strata:
\texttt{agent\_acted}, \texttt{ai\_as\_target}, \texttt{human\_acted\_on\_output}, and
\texttt{elicitation\_only}. Zero-click resource fetching counts as agent action when the system
makes the consequential request without an intervening human choice. We use four population
names throughout: the \emph{full catalog} contains all \N{} records; the
\emph{agent-acted subset} contains the \NagentSubset{} \texttt{agent\_acted} records; the
\emph{generative subset} contains the \NgenerativeSubset{} generative-system records; and the
\emph{primary population} is their \Nprimary{}-record intersection, comprising generative
systems in which the agent acted. Primary-population results support claims about agent
action; the other three populations show how those results change when either restriction is
relaxed.

\subsection{Corpus construction}

\begin{figure*}[t]
\centering
\includegraphics[width=\textwidth]{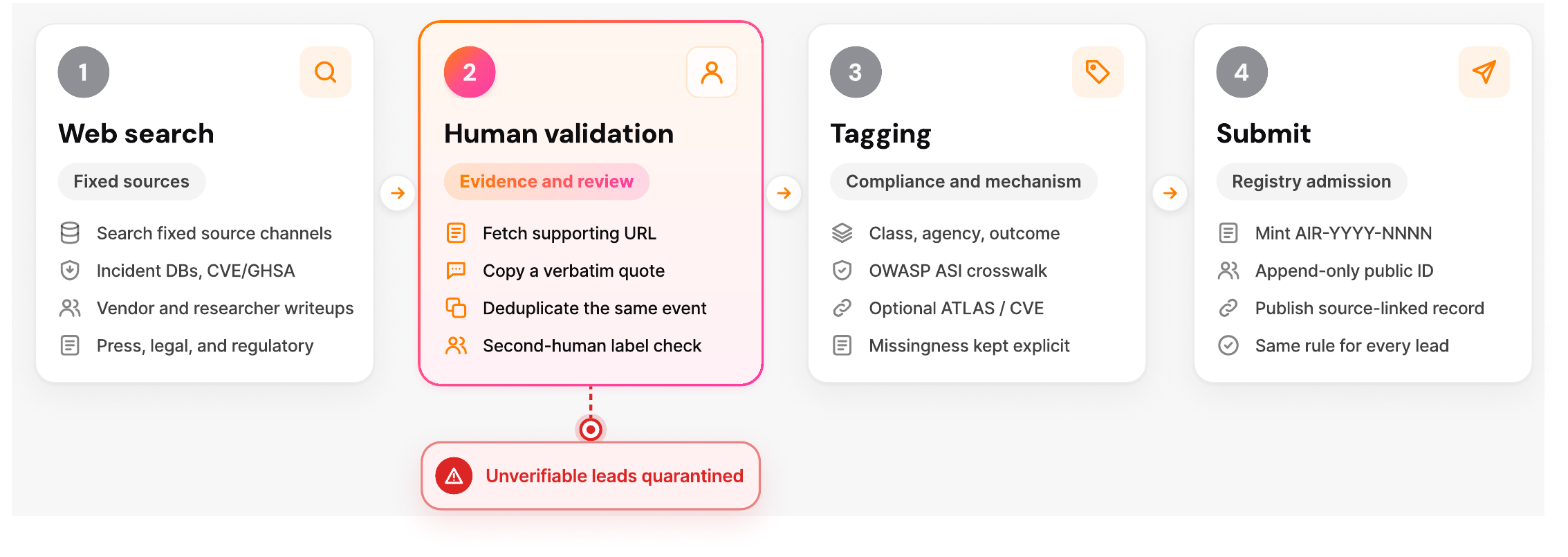}
\caption{AIR curation process. Each candidate is collected from a fixed source channel,
validated by a human against a fetched URL and verbatim quotation, tagged with mechanism
fields and compliance mappings (OWASP ASI, with optional ATLAS and CVE links), and
admitted under an append-only \texttt{AIR-YYYY-NNNN} identifier. Leads without fetchable
support are quarantined and never enter the catalog.}
\label{fig:pipeline}
\end{figure*}

\paragraph{Collection and deduplication.}
The same event or vulnerability often appears in several sources, so corpus construction had
two stages: collecting candidates and determining which described the same underlying
phenomenon. We first searched \Nshards{} source channels: incident databases; CVE/GHSA
advisories; vendor and independent-researcher disclosures; enterprise, coding-agent, browser,
MCP, and supply-chain reports; threat intelligence; academic demonstrations; and press, legal,
and regulatory records. These channels guided discovery but did not define the final labels;
together, they produced \Nshardrows{} candidates. Figure~\ref{fig:pipeline} shows the
recurring process that admits a record: search over those channels, human validation,
mechanism and compliance tagging, and registry submission.

We then linked candidates using exact identifiers, shared sources or quotations, vendor/date
matches, and title similarity. The curator reviewed every proposed link. Merging
\Nduplicates{} duplicate candidates yielded the \N{} records in the full catalog. Corpus counts
therefore do not rise merely because several sources describe the same event. The artifact
preserves each candidate-to-record mapping and merge decision so users can audit how the counts
were formed. Deduplication does not establish completeness: the source channels overlap and
were not randomly sampled, so duplicate reports cannot reveal how many events were never
disclosed or never found.

\paragraph{Evidence and adjudication.}
Admission required at least one fetched supporting URL and a verbatim quotation that establishes
the event or demonstrated vulnerability. We preferred a vendor, researcher, advisory, court, or
regulatory source. When the originating source was unavailable, a secondary report could support
a medium-confidence record; \Nhighconfidence{} records are high-confidence and
\Nmediumconfidence{} are medium-confidence. Leads without fetchable support were retained in a
separate quarantine (\Nquarantine{} at freeze) and do not enter any statistic.
These requirements establish source linkage and quote support, not independent reproduction
or proof by the authors. LLM assistance was limited to drafting the web-scraping and
figure-generation scripts. Human authors reviewed the code and its outputs; deterministic
validators enforced schema constraints, and the curator resolved scope, duplicate, and
classification decisions.

\subsection{Coding and reliability}

\paragraph{Outcome and disclosure class.}
\texttt{impact\_realized} is true only when a real party suffered a real consequence; a proof of
concept against a live product remains demonstrated capability. Separately, \texttt{class}
records how the event surfaced: an observed \texttt{in\_the\_wild} attack or attempt, a
\texttt{safety\_failure} with no adversary-supplied trigger, a shipped vulnerability or concrete
vendor threat mode surfaced through \texttt{responsible\_disclosure}, or a researcher-led
\texttt{research\_demo}.
Class does not entail outcome: an in-the-wild attempt or safety near miss can be unrealized, and
a controlled live-system test can still produce a real third-party consequence. Conditioning on
class diagnoses discovery composition; it is \emph{not} a causal adjustment.

\paragraph{Mechanism fields.}
Each record separates system context (attack surface, autonomy, capabilities, target, and
vendor), event pathway (initial vector, causal role, and consequences), response evidence
(guardrail description, bypass status, and remediation), and source metadata. Ambiguous
causal-role assignments carry a separate \texttt{agency\_debatable} flag. Controlled
derivations collapse free-text vectors into eight families and guardrail descriptions into
eleven multi-label kinds plus \texttt{none}/\texttt{unknown}. A versioned OWASP Agentic
Top~10 (ASI) crosswalk applies documented, mechanism-first rules to all \N{} records. All
\NasiInScope{} records within ASI's software-agent scope receive at least one code, while
\NasiOutOfScope{} records are explicitly marked \texttt{out\_of\_scope}; no in-scope record
remains unmapped. Existing curated codes retain their order, rule-supported codes are appended,
and reasoned overrides take precedence. The first ordered code becomes
\texttt{primary\_asi}, but records may carry multiple codes. ATLAS and CVE links remain
optional. Fields with inadequate coverage, including tool privilege and financial-loss bands,
are excluded from substantive analysis. Appendix~\ref{app:codebook} gives abbreviated decision
rules; the full codebook, crosswalk audit, and curation protocol accompany the artifact.

\paragraph{Human validation.}
One corpus curator made or adjudicated the initial labels. A second human reviewer then
inspected all \N{} records and their existing labels for completeness and correctness. Because
the reviewer could see those labels, this was not an independent annotation exercise. We
therefore do not report inter-rater agreement statistics.
The later rule-added ASI codes carry separate provenance and are not presented as independently
human-coded labels.
Remaining uncertainty is represented through explicit missing values, the
\texttt{agency\_debatable} flag, documented boundary rules, and sensitivity analyses.

\subsection{Analysis}
The analysis asks three questions in sequence. First, how do disclosure class and causal
role shape the realized-outcome shares in the frozen catalog? Second, how sensitive are those
shares to population definition and source concentration? Third, which observed mechanisms
does a selected agent evaluation represent, and which does it omit?

For each proportion, we report a 95\% Wilson interval and a deterministic pairs-cluster
bootstrap that resamples records sharing the same first-source host as a block (10,000
replicates; fixed seed). Both describe uncertainty within this disclosure sample; the
bootstrap captures one form of source dependence and is not a population confidence interval.
Records may also share a campaign or research program across hosts, so even that interval can
understate dependence. We therefore do not report record-independent hypothesis tests.
Autonomy is analyzed categorically through raw disclosure-class-by-autonomy cells; sparse
approval-gated cells preclude a precise adjusted estimate. Control outcomes are excluded from
quantitative analysis because their source coverage is inadequate. ASI assignments are
summarized only as non-exclusive crosswalk coverage, not as outcome, prevalence, or risk
estimates, because the mapping is partly rule-derived and the categories overlap.

\section{Measurements and evaluation audit}
\label{sec:findings}

The results proceed in two parts. We first establish how disclosure class, causal role,
population definition, and source concentration limit interpretation of outcome shares. We
then examine the apparent autonomy pattern and demonstrate AIR's intended evaluation-audit use
by comparing no-adversary failures with InjecAgent.

\subsection{Disclosure composition determines aggregate outcome shares}
The full catalog combines four disclosure classes and four causal roles. We report both
compositions before narrowing to the primary population.

\begin{table}[t]
\caption{Disclosure class and realized outcomes in the full \N{}-record catalog.
In-the-wild and safety-failure records are predominantly realized; responsible disclosures
and research demonstrations are predominantly demonstrated.}
\label{tab:composition}
\centering
\small
\setlength{\tabcolsep}{7pt}
\begin{tabular}{@{}lrr@{}}
\toprule
Disclosure class & $n$ & Realized, $n$ \\
\midrule
In the wild & \Nitw & \NitwRealized \\
Safety failure & \Nsafety & \NsafetyRealized \\
Responsible disclosure & \Ndisc & \NdiscRealized \\
Research demo & \Ndemo & \NdemoRealized \\
\bottomrule
\end{tabular}
\end{table}

\begin{table}[t]
\caption{Causal-role composition in the full \N{}-record catalog. Percentages are within role;
the $n=2$ \texttt{elicitation\_only} percentage is suppressed.}
\label{tab:agency}
\centering
\small
\setlength{\tabcolsep}{7pt}
\begin{tabular}{@{}lrr@{}}
\toprule
Agency stratum & $n$ & Realized, $n$ (\%) \\
\midrule
Agent acted & \NagentActed & \RagentActed{} (\PagentActed\%) \\
AI as target & \Ntarget & \Rtarget{} (\Ptarget\%) \\
Human acted on output & \NhumanActed & \RhumanActed{} (\PhumanActed\%) \\
Elicitation only & \Nelicitation & \Relicitation{} (\Pelicitation) \\
\bottomrule
\end{tabular}
\end{table}

Only \NclassOutcomeMismatch{} records depart from the pattern in which in-the-wild and
safety-failure records are realized while responsible disclosures and research demonstrations
are demonstrated. The latter two classes contribute \NresearchClass{} records but only
\Ndiscrealized{} realized cases; the former two contribute \NfieldClass{} records and
\NfieldClassRealized{} realized cases. The full catalog's realized share of
\Nrealized{}/\N{} (\Prealized\%) therefore primarily reflects the balance of disclosure
pathways, not the risk of deploying an agent. Table~\ref{tab:agency} separately shows why
causal role matters: target-side vulnerabilities and human action on model output should not
be counted as agent action.

Within the primary population, \Rprimary{} of \Nprimary{} records have realized harm
(\Pprimary\%; 95\% Wilson interval \Lprimary--\Uprimary\%). Resampling first-source hosts
widens the interval to \LprimaryCluster--\UprimaryCluster\%. Relaxing one restriction at a
time gives \RagentSubset{}/\NagentSubset{} (\PagentSubset\%) in the agent-acted subset and
\RgenerativeSubset{}/\NgenerativeSubset{} (\PgenerativeSubset\%) in the generative subset;
relaxing both gives the full catalog's value above. Restricting the primary population to
high-confidence records gives
\RprimaryHighConfidence{}/\NprimaryHighConfidence{} (\PprimaryHighConfidence\%).
These are disclosure-sample sensitivities, not estimates of failure incidence.

\begin{figure}[t]
\centering
\includegraphics[width=\linewidth]{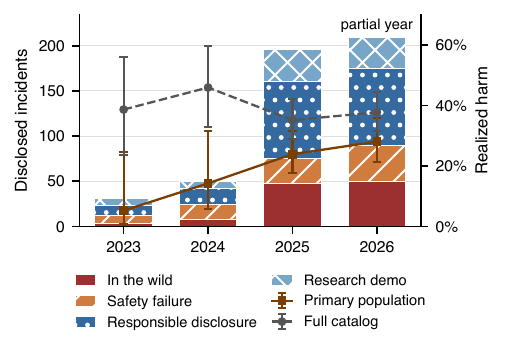}
\caption{Annual disclosure volume by class (bars) and realized-outcome shares with 95\%
Wilson intervals (points). The sole 2022 record is omitted from the plotted contrasts; 2026 is
right-truncated at \Freeze{}. Changes over time describe this catalog's disclosure composition,
not deployment risk.}
\label{fig:year-class}
\end{figure}

Figure~\ref{fig:year-class} makes the changing disclosure mix visible over time. Annual volume
and realized-outcome shares move with the balance of field events, disclosures, and
demonstrations. The figure is therefore a composition diagnostic, not a temporal failure-rate
or deployment-risk trend.

Agency-label uncertainty does not erase the agent-acted subset's result. Excluding all
\Ndebatable{} \texttt{agency\_debatable} records leaves \RagentStable{} realized cases among
\NagentStable{} uncontested agent-acted records (\PagentStable{}\%), compared with
\PagentActed{}\% under the coded labels. Reassigning all contested records in the directions
that minimize or maximize the share yields \PagentLowerBound{}\%--\PagentUpperBound{}\%.
These are identification bounds for the agent-acted subset, not sampling intervals.

\subsection{Source dependence dominates precision}
\begin{figure*}[t]
\centering
\includegraphics[width=\textwidth]{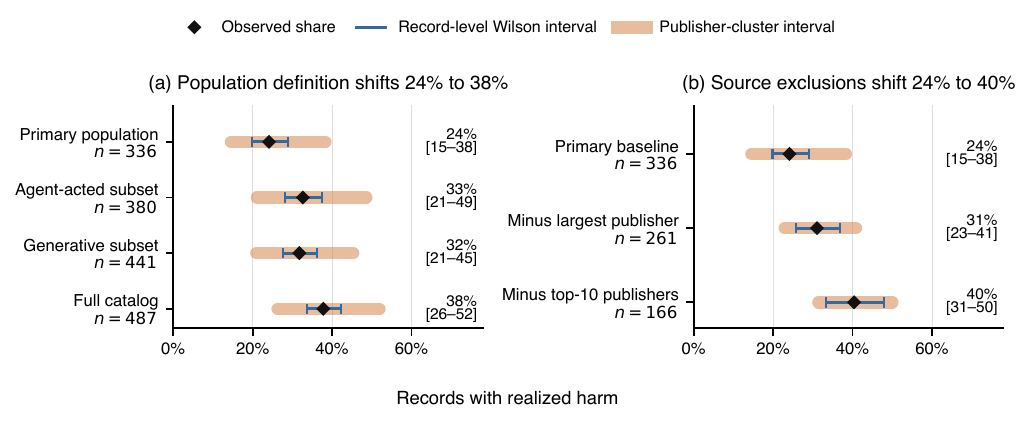}
\caption{Sensitivity of the realized-harm share. Panel (a) broadens the primary-population
definition; panel (b) removes dominant source blocks from the primary population. Diamonds are
within-sample estimates, thin blue intervals are record-level 95\% Wilson intervals, and wide
orange intervals are deterministic first-source-host bootstrap ranges. Labels report the point
estimate and the bracketed host-cluster range. These intervals characterize the selected
catalog, not deployment risk.}
\label{fig:sensitivity}
\end{figure*}

Publisher concentration matters: removing dominant source blocks moves the primary
population's \Pprimary\% realized share as high as \PprimaryWithoutTopTen\%. Removing each of
the ten largest first-source hosts in turn moves it between \PprimaryLosoMin{} and
\PprimaryLosoMax\%. Removing the largest source block, \texttt{embracethered.com}, leaves
\RprimaryWithoutEmbrace{}/\NprimaryWithoutEmbrace{} (\PprimaryWithoutEmbrace\%); removing all
ten largest hosts leaves \RprimaryWithoutTopTen{}/\NprimaryWithoutTopTen{}
(\PprimaryWithoutTopTen\%). Figure~\ref{fig:sensitivity} shows the four named populations and
separates population-definition changes from source-block exclusions. The perturbations
diagnose source dependence; they are not corrected estimates.

\subsection{Autonomy is confounded with disclosure class}
\begin{figure*}[t]
\centering
\includegraphics[width=\textwidth]{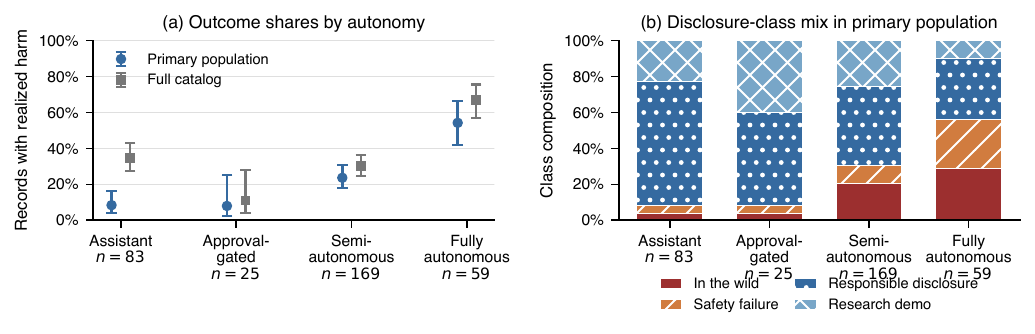}
\caption{Autonomy and disclosure composition. (a) Realized-harm shares in the primary
population and full catalog, with record-level 95\% Wilson intervals. (b) Disclosure classes
within the primary population. The apparent increase across autonomy labels coincides with a
composition shift and does not identify an autonomy effect.}
\label{fig:autonomy}
\end{figure*}

We next ask whether AIR supports a comparison across autonomy labels. Within the primary
population, realized-harm shares are \PassistPrimary\% for assistants,
\PcopilotPrimary\% for approval-gated copilots, \PsemiPrimary\% for semi-autonomous systems,
and \PfullPrimary\% for fully autonomous systems (Fig.~\ref{fig:autonomy}a).
This marginal pattern does not identify an autonomy effect: the corpus has neither deployment denominators
nor matched systems, and disclosure-class composition changes sharply across autonomy labels
(Fig.~\ref{fig:autonomy}b). The approval-gated group contains only one in-the-wild and one
safety-failure record, whereas the fully autonomous group contains 33 records across those
two classes.
Appendix~\ref{app:autonomy-cells} reports the underlying cells.

\subsection{No-adversary failures expose an evaluation gap}
The full catalog contains \Nsafety{} safety-failure records with no adversary-supplied
trigger, of which \NsafetyRealized{} are realized. The agent-acted subset contains
\NsafetyAgent{} of these records, including \NsafetyAgentRealized{} realized outcomes.
Documented mechanisms include destructive shell actions, uncommanded publication, and
embodied-control failures without adversarial goal hijacking. Evaluations limited to
attacker-supplied inputs do not exercise this region.

To make that comparison concrete, we project InjecAgent \cite{zhan2024injecagent} into AIR's
mechanism schema. Its \NInjecAgentCases{} base-setting cases combine
\NInjecAgentTools{} user tools with \NInjecAgentAttackCases{} attacker cases. Every task places
an attacker instruction in a user-tool response and therefore exercises indirect prompt
injection. The generic harness maps literally to \texttt{agent\_framework}. The reported
\NInjecAgentCoveredSurfaces{}-of-\NInjecAgentSurfaceUniverse{} statistic is a separate
deployment-analogue projection: it maps user tools to \texttt{coding\_agent},
\texttt{browser\_agent}, and \texttt{enterprise\_assistant}, and does not count the literal
harness as a fourth deployment surface.

\input{eval/injecagent-coverage.tex}

InjecAgent therefore does not test the no-adversary mechanisms identified above. Covering that
region would require workload, state, permission, and recovery perturbations without
attacker-supplied content. Disclosure class and realized outcome describe the public events
that motivate tests; they are not coverage dimensions that a synthetic task must itself
occupy. AIR record counts likewise do not determine benchmark priorities.

\section{Threats to validity}
\label{sec:threats}

\paragraph{Selection and denominator.}
AIR samples public disclosure, not deployed systems or agent runs. Counts track who publishes:
one independent research site supplies the largest source block, vendors with mature disclosure
programs become more visible, and controls that hold are seldom reported. Media attention,
regulation, product adoption, and researcher tooling all change over time. Consequently, no
count in this paper estimates incidence, prevalence, vendor risk, or control efficacy, and
class-stratified analysis remains descriptive rather than causal. The counting unit is also
nonuniform in scale: one coordinated advisory, malware campaign, or victim event receives one
record regardless of artifact, download, or victim count. Equal record weighting must not be
read as equal affected populations.

\paragraph{Evidence and annotation.}
A fetched quotation establishes that a source made a claim; it does not validate every detail of
that claim. \Nmediumconfidence{} records rely on secondary reporting. LLM assistance was
limited to drafting web-scraping and figure-generation code; human authors reviewed the code
and verified its outputs. One human curator made or adjudicated the initial labels, and a second
human reviewer inspected every record and its existing labels. This full-catalog review can
catch omissions and inconsistencies, but because the labels were visible it does not measure
independent coder agreement. The \texttt{agency\_debatable} flag exposes one known boundary and
the Results report its sensitivity, but other misclassification remains possible. Records
sharing a campaign, vendor, source, or research team remain dependent; resampling first-source
hosts captures only one of those links and can still understate uncertainty.

\paragraph{Geographic and temporal coverage.}
AIR does not code source language, incident country, or deployment geography, so this freeze
cannot quantify linguistic or regional representation. Collection used English-language search
and some non-English events rely on English-language AIID summaries; this is a plausible
coverage mechanism, not a measured geographic result. Disclosure dates have variable precision,
and 2026 is right-truncated at \Freeze{}.

\paragraph{System scope.}
The full catalog includes \NnonGenerative{} non-generative embodied-autonomy records, of which
\NrealizedNonGenerative{} are realized; the generative subset and primary population apply an
explicit \texttt{ai\_entity} filter. Two stable-ID boundary records are tagged
\texttt{not\_ai} rather than hidden. Excluding non-generative eligibility and ranking systems
also means AIR cannot support claims about automated discrimination generally.

\paragraph{Mechanism-field limits.}
Control outcomes are source-silent for \NguardSilent{} of \N{} records. Their descriptions
support case retrieval but not control-prevalence or efficacy claims. The ASI crosswalk has
complete in-scope coverage, but coverage is not independent validation: \NasiCurated{}
in-scope records retain at least one earlier curated assignment, \NasiRuleOnly{} are
rule-only, and \NasiOverrides{} use explicit overrides. Because assignments are non-exclusive
and partly derived from other labels, their counts support retrieval and evaluation-scope
auditing rather than category-prevalence or risk-ranking claims. Separately,
\NvectorOther{} vectors remain in the residual \texttt{other} family;
\texttt{tool\_access} remains unknown for all \NtoolAccessUnknown{} records; and authority,
reversibility, and financial-loss bands lack adequate source coverage.

\section{Artifact and use}
\label{sec:release}
The anonymized artifact accompanying this paper contains the frozen corpus, coding protocol,
versioned ASI crosswalk and audit, InjecAgent mapping, and code needed to reproduce every
reported statistic and figure.

\paragraph{Intended evaluation use.}
Evaluation designers can map test cases to AIR's surface, vector, and trigger fields, then
report represented and missing mechanisms. The realized/demonstrated distinction prevents
demonstrated capability from being counted as field harm. Because public disclosures lack
exposure denominators, AIR does not support attack-success rates, vendor rankings, or
deployment-risk scores \cite{gebru2021datasheets}.

\paragraph{Societal impact and safeguards.}
A shared, evidence-linked record can help evaluators and practitioners learn from failures that
would otherwise remain isolated. It can also amplify contested claims, simplify discovery of
exploit research, or invite unsupported vendor comparisons. AIR includes only short excerpts
needed to verify record admission, not full articles or new exploit payloads, and retains source
links, confidence labels, and correction provenance. These measures reduce but do not eliminate
reputational and dual-use risk.

\section{Conclusion}
AIR turns fragmented public reports into a source-linked record of how agentic systems fail.
Its central lesson is methodological: incident evidence can reveal which mechanisms an
evaluation represents only when realized harm, demonstrated capability, causal role, and
disclosure class remain separate. In the \Nprimary{}-record primary population, disclosure
class accounts descriptively for much of the apparent outcome and autonomy pattern; the
comparison with InjecAgent exposes a no-adversary gap in attack-only evaluations. Together,
these results show how selected disclosures can inform tests without being mistaken for
deployment rates. AIR provides a disciplined bridge from public failures to test design, with
source evidence and uncertainty attached.

\section*{Ethical Considerations}
AIR indexes claims from public sources that may identify vendors, researchers, organizations,
and individuals. An AIR identifier means that a report met the corpus evidence rule; it does
not independently confirm every claim in the source. The release contains only short evidence
excerpts and no new exploit payloads. An employer-independent correction process is planned but
was not operational at the corpus freeze. The authors' employer sells agent-security products,
creating a conflict of interest relevant to collection and coding decisions.

\section*{Open Science}
An anonymized artifact accompanying the submission contains the frozen corpus, analysis code,
human-validation description, and benchmark mapping needed to reproduce or audit the reported
results. Upon acceptance, we will archive this version under a persistent identifier. Release
terms will distinguish author-created annotations and code from third-party source excerpts.

\section*{LLM Usage Considerations}
For the research workflow, LLM assistance was limited to drafting the web-scraping and
figure-generation scripts. LLMs also assisted with language-level manuscript editing. Human
authors reviewed the code, outputs, prose, and citations. All scope, duplicate-resolution, and
labeling decisions were made or adjudicated by humans.

\bibliographystyle{plain}
\bibliography{refs}

\appendix
\section{Coding scheme}
\label{app:codebook}

\begin{table*}[t]
\caption{Core coding decisions used in the paper. Plain-language concepts are paired with
their schema fields; the complete attack-surface and vector vocabularies follow the table.}
\centering
\small
\setlength{\tabcolsep}{5pt}
\renewcommand{\arraystretch}{1.08}
\begin{tabular}{@{}>{\raggedright\arraybackslash}p{0.25\textwidth}
                    >{\raggedright\arraybackslash}p{0.70\textwidth}@{}}
\toprule
Concept and schema field & How it is coded \\
\midrule
\multicolumn{2}{@{}l}{\textit{Outcome, agency, and scope}} \\
\addlinespace[2pt]
\textbf{Disclosure route}\\[-1pt]\texttt{class} &
How the event became public: \emph{in the wild}, \emph{responsible disclosure},
\emph{research demonstration}, or \emph{safety failure}. Safety failure is reserved for
events without an adversary. \\
\addlinespace[3pt]
\textbf{Realized harm}\\[-1pt]\texttt{impact\_realized} &
\texttt{true} only when a real party experienced the reported consequence. A proof of
concept remains \texttt{false}, even on a live system, unless a real party was affected. \\
\addlinespace[3pt]
\textbf{Action autonomy}\\[-1pt]\texttt{autonomy\_level} &
The highest level of independent action shown in the event: \emph{assistant},
\emph{approval-gated copilot}, \emph{semi-autonomous}, or \emph{fully autonomous}. Product
marketing does not determine this label. \\
\addlinespace[3pt]
\textbf{Causal role}\\[-1pt]\texttt{agency} &
Whether the agent acted, the AI was the target, a human acted on AI output, or the source
only elicited a response. Borderline cases are marked \texttt{agency\_debatable}. \\
\addlinespace[3pt]
\textbf{AI type}\\[-1pt]\texttt{ai\_entity} &
Whether the relevant component is generative AI, non-generative autonomy, or not AI. This
field makes scope exclusions explicit. \\
\addlinespace[5pt]
\multicolumn{2}{@{}l}{\textit{Mechanism and retrieval}} \\
\addlinespace[2pt]
\textbf{Attack surface}\\[-1pt]\texttt{attack\_surface} &
The system surface directly implicated in the event. Multiple values may be assigned from
the 12-surface vocabulary listed below. \\
\addlinespace[3pt]
\textbf{Trigger pathway}\\[-1pt]\texttt{vector\_family} &
One of eight broad pathway families derived deterministically from the detailed
\texttt{initial\_vector}; the original wording is retained. \\
\addlinespace[3pt]
\textbf{Guardrail}\\[-1pt]\texttt{guardrail\_kind} &
Controls explicitly supported by the source. Multiple values are allowed; source silence is
\texttt{unknown}, not evidence that no guardrail existed. \\
\bottomrule
\end{tabular}
\end{table*}
\FloatBarrier

Three boundary examples operationalize autonomy. EchoLeak (\texttt{AIR-2025-0039}) is
\emph{assistant}: automatic email retrieval and rendering occurred during a user-initiated
Copilot interaction, although exploitation required no click on the malicious email; the
system did not own an ongoing task. Replit (\texttt{AIR-2025-0061}) is
\emph{semi-autonomous}: the agent issued several consequential commands without per-command
approval, but within a user-delegated coding task. The Remoteli posting bot
(\texttt{AIR-2022-0001}) is \emph{fully autonomous}: its persistent loop selected and
published text from monitored input without a contemporaneous human gate.
\emph{Approval-gated copilot} is reserved for configurations in which a consequential action
or tool connection crosses an explicit confirmation step. Coding follows authority exercised
in the event, not the product's advertised maximum.

The remaining controlled vocabularies used by the coverage audit are the surface and vector
families.
The twelve attack surfaces are \texttt{enterprise\_assistant}, \texttt{coding\_agent},
\texttt{browser\_agent}, \texttt{computer\_use\_agent}, \texttt{mcp\_server},
\texttt{agent\_framework}, \texttt{skill\_plugin}, \texttt{multi\_agent\_system},
\texttt{memory\_store}, \texttt{consumer\_chatbot}, \texttt{autonomous\_ops}, and
\texttt{other}. Vector families are indirect prompt injection, direct prompt injection,
malicious component, conventional vulnerability, misuse, physical environment, no-adversary
autonomous action, and other. The \texttt{safety\_failure} class identifies disclosures with
no adversary-supplied trigger; \texttt{no\_adversary\_autonomous\_action} separately identifies
the event's vector.

\subsection{Class-by-autonomy cells}
\label{app:autonomy-cells}
\begin{table*}[h]
\caption{Realized/total counts by disclosure class and autonomy in the primary population
(generative-system records in which the agent acted). Small approval-gated in-the-wild and
safety-failure cells preclude a precise adjusted estimate.}
\label{tab:class-autonomy}
\centering
\small
\setlength{\tabcolsep}{8pt}
\begin{tabular}{@{}lrrrr@{}}
\toprule
Disclosure class & Assistant & Approval-gated & Semi-autonomous & Fully autonomous \\
\midrule
In the wild & \RPrimaryItwAssist/\NPrimaryItwAssist & \RPrimaryItwCopilot/\NPrimaryItwCopilot &
  \RPrimaryItwSemi/\NPrimaryItwSemi & \RPrimaryItwFull/\NPrimaryItwFull \\
Safety failure & \RPrimarySafetyAssist/\NPrimarySafetyAssist & \RPrimarySafetyCopilot/\NPrimarySafetyCopilot &
  \RPrimarySafetySemi/\NPrimarySafetySemi & \RPrimarySafetyFull/\NPrimarySafetyFull \\
Responsible disclosure & \RPrimaryDiscAssist/\NPrimaryDiscAssist & \RPrimaryDiscCopilot/\NPrimaryDiscCopilot &
  \RPrimaryDiscSemi/\NPrimaryDiscSemi & \RPrimaryDiscFull/\NPrimaryDiscFull \\
Research demo & \RPrimaryDemoAssist/\NPrimaryDemoAssist & \RPrimaryDemoCopilot/\NPrimaryDemoCopilot &
  \RPrimaryDemoSemi/\NPrimaryDemoSemi & \RPrimaryDemoFull/\NPrimaryDemoFull \\
\bottomrule
\end{tabular}
\end{table*}

As a separate sensitivity in the full catalog, direct standardization of each autonomy group
to the full catalog's disclosure-class mix gives \PassistStandardized\%,
\PcopilotStandardized\%, \PsemiStandardized\%, and \PfullStandardized\%. The standard
population is itself disclosure-selected, and the one- and two-record approval-gated
in-the-wild and safety-failure cells do not support inferential adjustment.

\subsection{Boundary cases for realized outcomes}
Table~\ref{tab:realized-research} exposes all \Ndiscrealized{} records in the full catalog
that are both realized and coded as responsible disclosure or research demonstration. These
are narrow record-level judgments, not evidence that a demonstration implies population harm.

\begin{table*}[h]
\caption{All records in the full catalog coded both as realized and as responsible disclosure
or research demo. ``Basis'' gives the documented real-party consequence.}
\label{tab:realized-research}
\centering
\small
\setlength{\tabcolsep}{4pt}
\begin{tabular}{@{}>{\raggedright\arraybackslash}p{0.22\linewidth}
                    >{\raggedright\arraybackslash}p{0.17\linewidth}
                    >{\raggedright\arraybackslash}p{0.17\linewidth}
                    >{\raggedright\arraybackslash}p{0.33\linewidth}@{}}
\toprule
AIR ID & Class & Agency & Basis for realized label \\
\midrule
\texttt{AIR-2024-0014} & Research demo & Human acted on output & More than 30,000 authentic downloads and use in company repositories \\
\texttt{AIR-2025-0010} & Responsible disclosure & Agent acted & Private repository contents and live secrets returned from Copilot's cache \\
\texttt{AIR-2025-0012} & Research demo & AI as target & 11,908 credentials authenticated; affected vendors rotated or revoked keys \\
\texttt{AIR-2026-0018} & Research demo & Agent acted & Attacker-controlled code executed on 16 non-consenting users' machines \\
\texttt{AIR-2026-0036} & Responsible disclosure & Agent acted & Exposed release token was used to publish an unauthorized package version \\
\bottomrule
\end{tabular}
\end{table*}

\subsection{OWASP Agentic Top 10 crosswalk}
\begin{figure}[t]
\centering
\includegraphics[width=\linewidth]{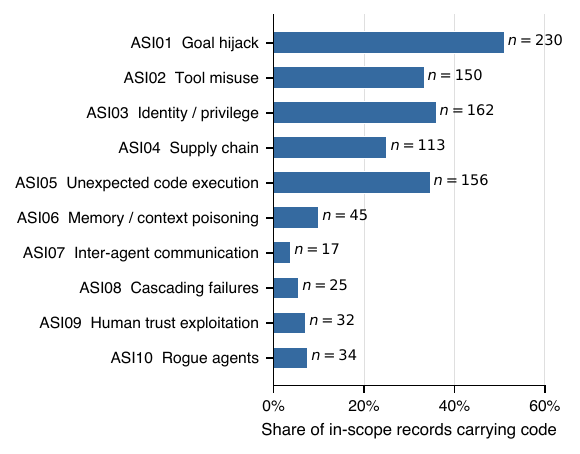}
\caption{Non-exclusive OWASP Agentic Top~10 assignments among \NasiInScope{} in-scope
records. Bars show the share carrying each code and labels give counts. Because records can
carry multiple mechanisms, bars do not sum to 100\%. The \NasiOutOfScope{} out-of-scope records
are excluded; no in-scope record is unmapped.}
\label{fig:asi}
\end{figure}

\paragraph{Purpose and coverage.}
AIR uses the OWASP Agentic Top~10 as a searchable index of documented mechanisms, not as a
severity score or an estimate of real-world risk. Of the full catalog, all \NasiInScope{}
records within OWASP's agentic-application scope have at least one code. The other
\NasiOutOfScope{} records---primarily non-generative perception and planning failures---are
marked \texttt{out\_of\_scope} rather than left blank. Codes are non-exclusive:
\NasiLabels{} assignments cover the in-scope records (\AsiLabelsPerRecord{} per record on
average), and \NasiMulti{} records carry more than one. Figure~\ref{fig:asi} shows the
distribution; Table~\ref{tab:asi} gives the evidence required for each code.

\paragraph{How assignments are made.}
Each record stores its dominant code first as \texttt{primary\_asi}. Among in-scope records,
\NasiCurated{} retain at least one earlier curator label, \NasiRuleOnly{} are rule-only, and
\NasiOverrides{} use an explicit override. Rules use structured fields wherever possible and
consult titles or notes only for evidence those fields do not capture. AIR stores the reason
and provenance for every mapping; the generated audit lists additions, overrides, and
differences between rule-based ordering and the curator's primary choice. Thus the crosswalk is
reviewable and reproducible, while its counts remain descriptions of this catalog rather than
estimates of mechanism prevalence or comparative risk.

\begin{table*}[t]
\caption{Decision rules for the non-exclusive OWASP Agentic Top~10 crosswalk. A code is
assigned only when the record's source evidence supports the stated mechanism.}
\label{tab:asi}
\centering
\small
\setlength{\tabcolsep}{3pt}
\begin{tabular}{@{}l>{\raggedright\arraybackslash}p{0.22\textwidth}
                    >{\raggedright\arraybackslash}p{0.68\textwidth}@{}}
\toprule
Code & Name & Minimum source-supported evidence \\
\midrule
ASI01 & Agent Goal Hijack & Attacker-supplied instructions or content redirect the agent's
objective, decisions, or action path. \\
ASI02 & Tool Misuse and Exploitation & The agent uses a legitimate tool unsafely while
remaining within its granted privileges. \\
ASI03 & Identity and Privilege Abuse & Delegation, credentials, authorization, identity, or
inherited privileges are abused. \\
ASI04 & Agentic Supply Chain Vulnerabilities & A skill, plugin, MCP server, package, model,
dataset, registry, or update is hostile, compromised, or tampered with. \\
ASI05 & Unexpected Code Execution & Code or command execution, unsafe deserialization, or a
sandbox escape reaches a path that should not have been executable. \\
ASI06 & Memory and Context Poisoning & Stored or retrievable context is corrupted and affects
later reasoning, planning, or tool use. \\
ASI07 & Insecure Inter-Agent Communication & An in-flight message between agents is injected,
spoofed, replayed, intercepted, or altered. \\
ASI08 & Cascading Failures & A fault propagates beyond its origin across agents, sessions,
workflows, or a fleet. \\
ASI09 & Human-Agent Trust Exploitation & A human over-relies on deceptive or incorrect agent
output and takes the consequential action. \\
ASI10 & Rogue Agents & An agent behaves harmfully or deceptively outside its intended function
or authorized scope, including adversary-operated agent use. \\
\bottomrule
\end{tabular}
\end{table*}
\FloatBarrier

\section{Dataset documentation}
\paragraph{Composition and provenance.}
The release contains public-source incident metadata and short verbatim evidence quotes; it does
not redistribute full articles. Records can name vendors, researchers, affected organizations,
and people already named in public reporting. Users should follow the linked source for context
and corrections.

\paragraph{Maintenance and corrections.}
New records receive new IDs. Existing records are corrected in place with provenance. When
records are merged, superseded references become aliases; IDs are never reused. Quarantined
leads are excluded until a supporting source is fetchable. The event set analyzed here was
frozen on \Freeze{}. The live registry is reviewed weekly for new records, source availability,
and corrections. Updates after the freeze are versioned separately and must not be substituted
when reproducing these results.

\paragraph{Uses and risks.}
Intended uses are evaluation-coverage audits, case retrieval, qualitative failure analysis, and
descriptive study of disclosed incidents. Unsupported uses include vendor league tables,
prevalence estimates, causal claims about autonomy or controls, and automated severity decisions.
Quotes may repeat allegations or descriptions of harm; citation of an AIR record should not be
read as endorsement of every source claim.

\end{document}

%% file: numbers.tex
\newcommand{\N}{487}
\newcommand{\Nrealized}{184}
\newcommand{\Prealized}{38}

\newcommand{\Nprimary}{336}
\newcommand{\Rprimary}{81}
\newcommand{\Pprimary}{24}
\newcommand{\Lprimary}{20}
\newcommand{\Uprimary}{29}
\newcommand{\LprimaryCluster}{15}
\newcommand{\UprimaryCluster}{38}
\newcommand{\NprimaryHighConfidence}{314}
\newcommand{\RprimaryHighConfidence}{70}
\newcommand{\PprimaryHighConfidence}{22}
\newcommand{\NclassOutcomeMismatch}{28}

\newcommand{\Nhighconfidence}{457}
\newcommand{\Nmediumconfidence}{30}

\newcommand{\NasiInScope}{451}
\newcommand{\NasiOutOfScope}{36}

\newcommand{\NasiCurated}{271}
\newcommand{\NasiRuleOnly}{175}
\newcommand{\NasiOverrides}{5}
\newcommand{\NasiLabels}{964}
\newcommand{\NasiMulti}{350}

\newcommand{\AsiLabelsPerRecord}{2.14}

\newcommand{\Nshards}{17}
\newcommand{\Nshardrows}{602}
\newcommand{\Nquarantine}{52}
\newcommand{\Nduplicates}{115}
\newcommand{\Nitw}{110}

\newcommand{\Nsafety}{92}

\newcommand{\NsafetyAgent}{82}
\newcommand{\Ndisc}{199}

\newcommand{\Ndemo}{86}

\newcommand{\NfieldClass}{202}
\newcommand{\NresearchClass}{285}
\newcommand{\NfieldClassRealized}{179}
\newcommand{\NitwRealized}{92}
\newcommand{\NsafetyRealized}{87}
\newcommand{\NsafetyAgentRealized}{77}
\newcommand{\NdiscRealized}{2}
\newcommand{\NdemoRealized}{3}

\newcommand{\Ndiscrealized}{5}

\newcommand{\NguardSilent}{390}

\newcommand{\NtoolAccessUnknown}{487}

\newcommand{\PfullPrimary}{54}

\newcommand{\PsemiPrimary}{24}

\newcommand{\PassistPrimary}{8}

\newcommand{\PcopilotPrimary}{8}
\newcommand{\PassistStandardized}{42}
\newcommand{\PcopilotStandardized}{41}
\newcommand{\PsemiStandardized}{34}
\newcommand{\PfullStandardized}{45}

\newcommand{\Ndebatable}{113}

\newcommand{\NagentStable}{315}
\newcommand{\RagentStable}{100}
\newcommand{\PagentStable}{32}
\newcommand{\PagentLowerBound}{27}
\newcommand{\PagentUpperBound}{42}
\newcommand{\NagentActed}{380}
\newcommand{\RagentActed}{124}
\newcommand{\PagentActed}{33}
\newcommand{\Ntarget}{69}
\newcommand{\Rtarget}{29}
\newcommand{\Ptarget}{42}
\newcommand{\NhumanActed}{36}
\newcommand{\RhumanActed}{31}
\newcommand{\PhumanActed}{86}
\newcommand{\Nelicitation}{2}
\newcommand{\Relicitation}{0}
\newcommand{\Pelicitation}{\textemdash}

\newcommand{\NnonGenerative}{44}

\newcommand{\NrealizedNonGenerative}{43}

\newcommand{\NagentSubset}{380}
\newcommand{\RagentSubset}{124}
\newcommand{\PagentSubset}{33}

\newcommand{\NgenerativeSubset}{441}
\newcommand{\RgenerativeSubset}{140}
\newcommand{\PgenerativeSubset}{32}

\newcommand{\NprimaryWithoutEmbrace}{261}
\newcommand{\RprimaryWithoutEmbrace}{81}
\newcommand{\PprimaryWithoutEmbrace}{31}
\newcommand{\NprimaryWithoutTopTen}{166}
\newcommand{\RprimaryWithoutTopTen}{67}
\newcommand{\PprimaryWithoutTopTen}{40}

\newcommand{\PprimaryLosoMin}{23}
\newcommand{\PprimaryLosoMax}{31}

\newcommand{\Yfirst}{2022}
\newcommand{\Ylast}{2026}
\newcommand{\Freeze}{2026-09-05}

\newcommand{\NvectorOther}{113}

\newcommand{\NPrimaryItwCopilot}{1}
\newcommand{\RPrimaryItwCopilot}{1}

\newcommand{\NPrimaryItwAssist}{3}
\newcommand{\RPrimaryItwAssist}{2}

\newcommand{\NPrimaryItwSemi}{35}
\newcommand{\RPrimaryItwSemi}{24}

\newcommand{\NPrimaryItwFull}{17}
\newcommand{\RPrimaryItwFull}{17}

\newcommand{\NPrimarySafetyCopilot}{1}
\newcommand{\RPrimarySafetyCopilot}{1}

\newcommand{\NPrimarySafetyAssist}{4}
\newcommand{\RPrimarySafetyAssist}{4}

\newcommand{\NPrimarySafetySemi}{17}
\newcommand{\RPrimarySafetySemi}{16}

\newcommand{\NPrimarySafetyFull}{16}
\newcommand{\RPrimarySafetyFull}{13}

\newcommand{\NPrimaryDiscCopilot}{13}
\newcommand{\RPrimaryDiscCopilot}{0}

\newcommand{\NPrimaryDiscAssist}{57}
\newcommand{\RPrimaryDiscAssist}{1}

\newcommand{\NPrimaryDiscSemi}{74}
\newcommand{\RPrimaryDiscSemi}{0}

\newcommand{\NPrimaryDiscFull}{20}
\newcommand{\RPrimaryDiscFull}{1}

\newcommand{\NPrimaryDemoCopilot}{10}
\newcommand{\RPrimaryDemoCopilot}{0}

\newcommand{\NPrimaryDemoAssist}{19}
\newcommand{\RPrimaryDemoAssist}{0}

\newcommand{\NPrimaryDemoSemi}{43}
\newcommand{\RPrimaryDemoSemi}{0}

\newcommand{\NPrimaryDemoFull}{6}
\newcommand{\RPrimaryDemoFull}{1}

\newcommand{\Ny}[1]{\csname Ny@#1\endcsname}
\newcommand{\Ry}[1]{\csname Ry@#1\endcsname}
\newcommand{\Py}[1]{\csname Py@#1\endcsname}
\newcommand{\Ly}[1]{\csname Ly@#1\endcsname}
\newcommand{\Uy}[1]{\csname Uy@#1\endcsname}
\newcommand{\NyAgent}[1]{\csname NyAgent@#1\endcsname}
\newcommand{\RyAgent}[1]{\csname RyAgent@#1\endcsname}
\newcommand{\PyAgent}[1]{\csname PyAgent@#1\endcsname}
\newcommand{\NyPrimary}[1]{\csname NyPrimary@#1\endcsname}
\newcommand{\RyPrimary}[1]{\csname RyPrimary@#1\endcsname}
\newcommand{\PyPrimary}[1]{\csname PyPrimary@#1\endcsname}
\expandafter\def\csname Ny@2022\endcsname{1}
\expandafter\def\csname Ry@2022\endcsname{1}
\expandafter\def\csname Py@2022\endcsname{\textemdash}
\expandafter\def\csname Ly@2022\endcsname{\textemdash}
\expandafter\def\csname Uy@2022\endcsname{\textemdash}
\expandafter\def\csname NyAgent@2022\endcsname{1}
\expandafter\def\csname RyAgent@2022\endcsname{1}
\expandafter\def\csname PyAgent@2022\endcsname{\textemdash}
\expandafter\def\csname NyPrimary@2022\endcsname{1}
\expandafter\def\csname RyPrimary@2022\endcsname{1}
\expandafter\def\csname PyPrimary@2022\endcsname{\textemdash}
\expandafter\def\csname Ny@2023\endcsname{31}
\expandafter\def\csname Ry@2023\endcsname{12}
\expandafter\def\csname Py@2023\endcsname{39}
\expandafter\def\csname Ly@2023\endcsname{24}
\expandafter\def\csname Uy@2023\endcsname{56}
\expandafter\def\csname NyAgent@2023\endcsname{28}
\expandafter\def\csname RyAgent@2023\endcsname{10}
\expandafter\def\csname PyAgent@2023\endcsname{36}
\expandafter\def\csname NyPrimary@2023\endcsname{19}
\expandafter\def\csname RyPrimary@2023\endcsname{1}
\expandafter\def\csname PyPrimary@2023\endcsname{5}
\expandafter\def\csname Ny@2024\endcsname{50}
\expandafter\def\csname Ry@2024\endcsname{23}
\expandafter\def\csname Py@2024\endcsname{46}
\expandafter\def\csname Ly@2024\endcsname{33}
\expandafter\def\csname Uy@2024\endcsname{60}
\expandafter\def\csname NyAgent@2024\endcsname{38}
\expandafter\def\csname RyAgent@2024\endcsname{14}
\expandafter\def\csname PyAgent@2024\endcsname{37}
\expandafter\def\csname NyPrimary@2024\endcsname{28}
\expandafter\def\csname RyPrimary@2024\endcsname{4}
\expandafter\def\csname PyPrimary@2024\endcsname{14}
\expandafter\def\csname Ny@2025\endcsname{196}
\expandafter\def\csname Ry@2025\endcsname{69}
\expandafter\def\csname Py@2025\endcsname{35}
\expandafter\def\csname Ly@2025\endcsname{29}
\expandafter\def\csname Uy@2025\endcsname{42}
\expandafter\def\csname NyAgent@2025\endcsname{152}
\expandafter\def\csname RyAgent@2025\endcsname{43}
\expandafter\def\csname PyAgent@2025\endcsname{28}
\expandafter\def\csname NyPrimary@2025\endcsname{142}
\expandafter\def\csname RyPrimary@2025\endcsname{34}
\expandafter\def\csname PyPrimary@2025\endcsname{24}
\expandafter\def\csname Ny@2026\endcsname{209}
\expandafter\def\csname Ry@2026\endcsname{79}
\expandafter\def\csname Py@2026\endcsname{38}
\expandafter\def\csname Ly@2026\endcsname{32}
\expandafter\def\csname Uy@2026\endcsname{45}
\expandafter\def\csname NyAgent@2026\endcsname{161}
\expandafter\def\csname RyAgent@2026\endcsname{56}
\expandafter\def\csname PyAgent@2026\endcsname{35}
\expandafter\def\csname NyPrimary@2026\endcsname{146}
\expandafter\def\csname RyPrimary@2026\endcsname{41}
\expandafter\def\csname PyPrimary@2026\endcsname{28}

%% file: eval/injecagent-numbers.tex
\newcommand{\NInjecAgentCases}{1,054}
\newcommand{\NInjecAgentTools}{17}
\newcommand{\NInjecAgentAttackCases}{62}
\newcommand{\NInjecAgentCoveredSurfaces}{3}
\newcommand{\NInjecAgentSurfaceUniverse}{12}

%% file: eval/injecagent-coverage.tex
\begin{table}[t]
\caption{Under the deployment-analogue projection, InjecAgent's \NInjecAgentCases{} cases
occupy three of AIR's twelve surfaces. The literal generic harness is not counted as a fourth
surface; every case is attacker-triggered indirect prompt injection.}
\label{tab:injecagent}
\centering
\small
\setlength{\tabcolsep}{8pt}
\begin{tabular}{@{}lrr@{}}
\toprule
Surface analogue & Cases & Share \\
\midrule
Enterprise assistant & 434 & 41.2\% \\
Browser agent & 434 & 41.2\% \\
Coding agent & 186 & 17.6\% \\
Other deployment surfaces & 0 & 0.0\% \\
\bottomrule
\end{tabular}
\end{table}